# Quantum Inspired High Dimensional Conceptual Space as KID Model for Elderly Assistance

Ishwarya M S, Aswani Kumar Ch.

School of Information Technology and Engineering,
Vellore Institute of Technology, Vellore.
Email: cherukuri@acm.org

**Abstract.** In this paper, we propose a cognitive system that acquires knowledge on elderly daily activities to ensure their wellness in a smart home using a Knowledge-Information-Data (KID) model. The novel cognitive framework called high dimensional conceptual space is proposed and used as KID model. This KID model is built using geometrical framework of conceptual spaces and formal concept analysis (FCA) to overcome imprecise concept notation of conceptual space with the help of topology based FCA. By doing so, conceptual space can be represented using Hilbert space. This high dimensional conceptual space is quantum inspired in terms of its concept representation. The knowledge learnt by the KID model recognizes the daily activities of the elderly. Consequently, the model identifies the scenario on which the wellness of the elderly has to be ensured.

## 1 Introduction

Health decline is an increasing problem in elderly people leading to the need for special attention. With the current advancements in robotics and wearable technology, feasible devices are developed for elderly care [1]. However, most of proposed solutions are expensive, awkward and troublesome and thereby creating a cognitive health decline [2]. One possible solution of monitoring elderly wellness without disturbing daily activities is Active Assisted Living using IOT [3]. In this regard, data from smart home is collected for analysis and action at critical time. Using this acquired data, activities of the elderly are analyzed via activity recognition techniques [4]. Inspired by human cognition process, literature reveals that human activities can be represented as concepts in cognitive system [3]. The models that transform information acquired from the real world data to knowledge are called as Knowledge-Information-Data (KID) model [5]. Any model that could perceives data, extract information, builds knowledge just like humans learning process can be regarded as KID models [6].

Various cognitive frameworks have been proposed to represent knowledge and treat uncertainty (inadequate data) [7]. Among those, our particular interest is on conceptual spaces and FCA as they both can model human activity recognition [3], [8]–[10]. However, notion of concept is imprecise in the conceptual spaces framework requiring topological definition while topology oriented FCA is limited to classical probability theory.

In this paper, we propose cognitive system to assist elderly by adapting a novel KID model called high dimensional conceptual space that has been built using both geometrical framework of conceptual space and FCA. By this treatment, the modified conceptual space framework called high dimensional conceptual space is represented in Hilbert space and shows quantum inspiration [11]–[13]. The proposed cognitive system alerts the kin of the elderly to ensure their wellness when the adapted KID attains uncertainty that is not resolved with time. The rest of the paper is organized as follows. Section 2 of the paper discusses a short literature review and in section 3 we have provided our detailed derivation of the proposed high dimensional conceptual space with its quantum inspiration. Section 4 discusses the proposed cognitive system to assist elderly and the successive section elaborates on experimental analysis.

## 2    Literature review

Cognitive frameworks can well be regarded as KID model since they provide a pragmatic transformation of data to knowledge in the form of concepts [5]. Concepts are learnt based on the cognitive contextual data. Generally, a context possess information relating to 'what', 'where', 'when' , 'with' and 'who' against 'why' [14]. Uncertainty in a cognitive system can be defined as absence or inadequate information that is required to perform reasoning. Conceptual spaces, FCA and relation algebra are cognitive frameworks which can used for modeling human activity recognition [10], [15], [16]. Wang et. al suggests human activity can be represented as relational concepts [16]. According to Gärdenfors, actions can be represented using geometrical framework of conceptual space by adding a force pattern [10]. However, he suggests that conceptual space framework has imprecise notation of concepts and requires a topology based definition. FCA is a mathematics based framework that acquires knowledge in the form of concepts from a context consisting objects described by their attribute [17]–[20]. FCA can be modeled using associative memories [21]. Literature reveals that associative memories are treated as a KID model and can recognize human activity [3]. Therefore, FCA can well be used for modeling activity recognition considering the ability of associative memories for modeling activity recognition. On the other hand, machine learning algorithms such as artificial neural network (ANN), support vector machines (SVM) and hidden markov model (HMM) can also be used for recognizing daily activities of elderly [22]. Cognitive vision system understands the visual information in the environment by integrating knowledge acquired from environment [23]. One potential application of cognitive vision system is human activity recognition. Adding to these, mining algorithms such as ProM tool and cognitive robots are also used for recognizing human activities [24]. In order to detect the health decline in elderly, it is necessary to detect the anomalies or abnormal behavior in their daily activities [25]. Uncertainty is an anomaly in a cognitive system. Based on these observations, in this paper, we propose a cognitive system by adapting a novel KID model for assisting elderly by recognizing their daily activities. Following this brief review on literature, we propose a novel KID model called high dimensional conceptual space in the next section of the paper.

## 3  High dimensional conceptual space

Let R be the smart home scenario in which we wish to assist elderly and E be the set of all activities of their daily living. According to geometrical framework of conceptual spaces, an event $e_i \subset E$ in R is represented by set of quality dimensions D such that $D = \sum_{j=1}^{n} d_j$ [8]. Each $d_j \subset D$ portrays a group of attributes of the corresponding quality dimension required to accomplish event $e_i$ [10]. A set of separable and interrelated quality dimensions forms a domain. The set of quality dimensions under different domains form a region in a conceptual space and this region is called a concept in geometrical framework of conceptual spaces. According to Gärdenfors, this notion of concept is imprecise with just quality dimensions and without topological definition [8]. To overcome this, we see through the notions of concepts in other cognitive frameworks for topology oriented definition [7]. Our particular interest is on FCA since it is topology oriented and it has connections with human cognitive processes[26]. According to FCA, a concept is pair consisting of objects (extent) and its associated set of attributes (intent). Three-way FCA is special kind of FCA, in which concept is a triple represented by set of objects, contributing set of attributes and non-contributing set of attributes [27]. Let H be the proposed high dimensional conceptual space in which we represent a concept's extent (event x) in terms of its intent consisting of contributing as well non-contributing attributes under different quality dimensions as shown equation 1.

$$x = \sum_{i=1}^{n} \sum_{j=1}^{m} d_i a_j I_j \qquad (1)$$

We regard that an attribute $a_j$ as a contributing attribute based on the binary relation $I_j$, i.e. if $I_j$ is 1 (0), then the corresponding attribute $a_j$ is contributing (non-contributing). The proposed high dimensional conceptual adapts the redefined concept denotation along with the geometrical framework of conceptual spaces.

In order to obtain the quantum inspiration behind the proposed high dimensional conceptual space, let us compare it with a quantum system. Let us consider a *N* qubit quantum system *S* formed by the set of basis vectors namely $|\phi_1>, |\phi_2>, ... |\phi_n>$ whose corresponding probability amplitudes are $\alpha_1 \alpha_2 ... \alpha_n$. A quantum state $|\psi>$ in *S* is represented by linear superposition of its basis vectors as shown in equation 2.

$$|\psi> = \sum_{k=1}^{N} \alpha_k |\phi_k> \qquad (2)$$

In the above equation, $\alpha_k$ represents the probability amplitude of the projection $|\phi_k>$. Let us compare term by term of equation 1 and 2 for quantum inspiration behind *H*. The term $a_j I_j$ is the projection corresponding to different attributes under a quality dimension $d_i$ and it is column vector. This projection vector $a_j I_j$ in equation 1 is analogous to the basis vector $|\phi_k>$ in equation 2. Similarly, the probability amplitude (angle) $\alpha_i$ of projection determines its importance to obtain the quantum state $|\psi>$. The probability amplitude term $\alpha_i$ is analogous to the quality dimension $d_i$ which portrays the qualities of the concept (event). Inspired by this analogy, we propose the quantum inspiration behind the high dimensional conceptual space $H$ as follows: 1) A concept in a high dimensional conceptual space is equivalent to a state in a quantum system. 2) An attribute in a high dimensional conceptual space is equivalent to the basis state in quantum system. 3) Relation between the attributes of a quality dimension with an event is equivalent to projection of a basis vector. In the following, we propose a formal method to assist elderly by recognition their daily activities.

## 4    Proposed Work

The proposed model adapts the high dimensional conceptual space as the KID model to assist elderly in a smart home. It is assumed that the sensors are equipped wherever necessary and useful in a smart home and these sensor data are centrally collected appropriately from the smart home. A set of time-driven sensor data that describes all possible actions or actions of interest is pre-processed and its corresponding context is created. We regard the aforementioned set of time-driven sensor data as training sensor data. This training sensor data is fed to KID model. Each activity of the elderly is learnt in the form of concepts. We regard the high dimensional conceptual space as KID model since it processes information obtained from the incoming data to acquire knowledge. Each action is represented by its sensor (quality dimensions), sub-type sensor data (attributes), and its relation with the action. Consequently, a high dimensional conceptual space (memory) is created consisting of concepts as geometrical structures formed by vector of associated quality dimensions and its attributes. Upon learning the list the concepts based on the training sensor data, the high dimensional conceptual space can now assist elderly based on real time sensor data acquired from the smart home. As represented Figure 1, the assistance model is cyclic evolutionary model. This nature of model as well as modeling element (action) introduces a time space upon initial learning.

Let $A$ be the set of time driven activities that are learnt by KID model from the training sensor data such that $A = a_1, a_2, \ldots a_n$ where $n$ is the number of activities learnt. Let $E$ be the time taken by each activity such that $E = e_1, e_2, \ldots e_n$. Let $T$ be reference time and it is set to maximum time taken by longest activity in training sensor data. With this initial setup, a real time sensor data is given as input to the conceptual space, i.e. pieces of description cue corresponding to a single action are received at different time granules $\Delta T$, such that $\Delta T = 1,2,\ldots T$. Upon receiving each piece of the description cue, the proposed model ensures the certainty of the KID model. We regard the model is certain if the model possess a matching geometric structure (associated vector of quality dimensions and attributes) similar to description cue in its conceptual space $\Delta C$ at time $\Delta T$ else the model is uncertain (Arecchi,2018). The conceptual space (memory) $\Delta C$ coupled with time space $\Delta T$ produces a Heisenberg's like uncertainty. For a given pieces of description cue, the proposed model can take three different ways to assist elderly.

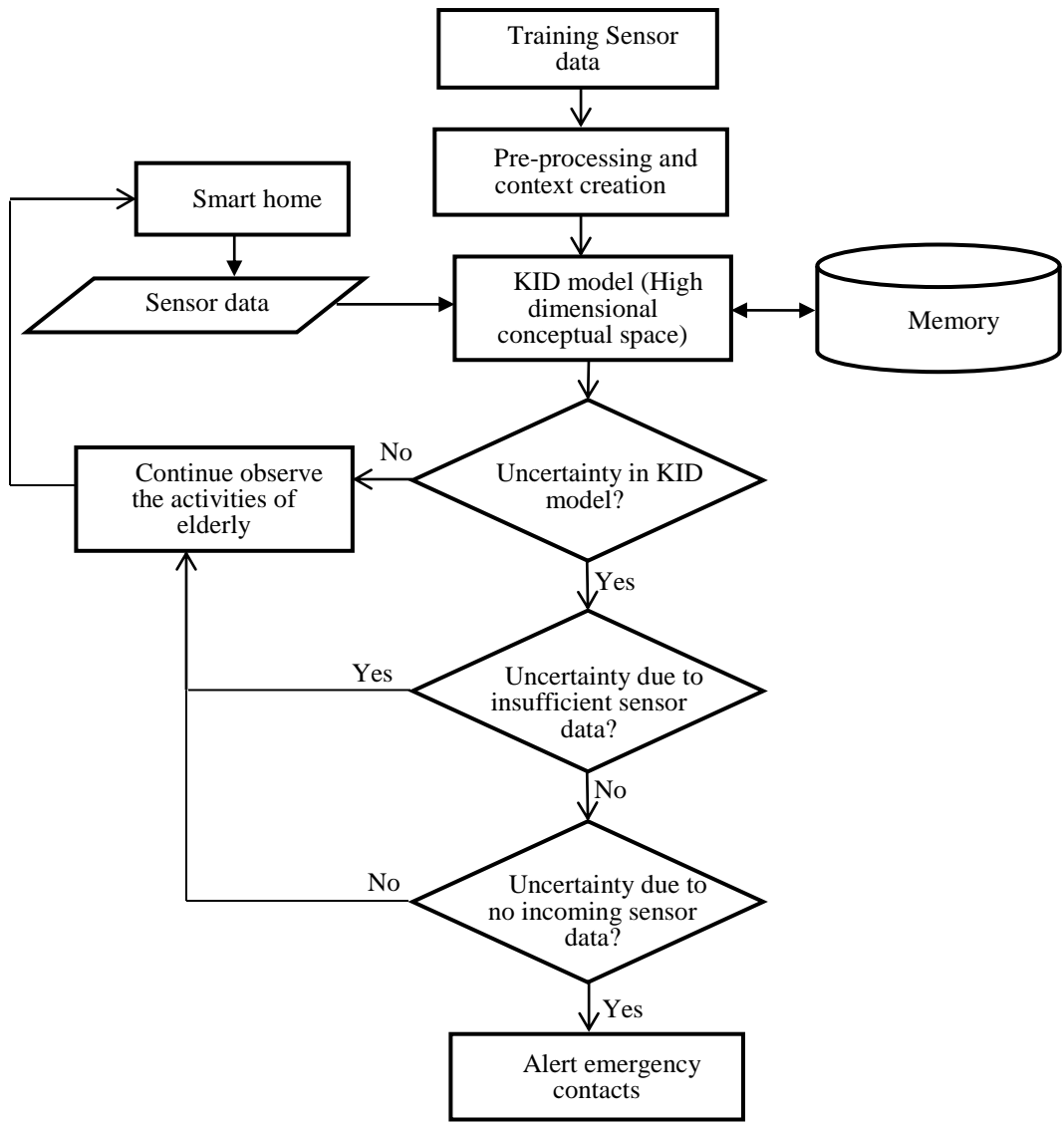

**Fig. 1.** High dimensional conceptual space as KID model for elderly assistance

1. No uncertainty: If $\Delta T < T$ and $\Delta C$ have a geometrical projection equivalent to the given description cue, then the model is certain at time $\Delta T$. The recognized elderly activity is the activity corresponding to the aforementioned geometrical projection. This condition indicates that the elderly is currently pursuing an activity in wellness which is still need to be completed. In this case, the proposed model adjusts the reference time $\Delta T$ in accordance to the most similar description in $\Delta C$ until the time $\Delta T$ (Muthukrishnan, 2006). Based on the aforementioned literature, the proposed model regards the smallest superset corresponding to the given cue until time $\Delta T$ as the most similar description in $\Delta C$. If there are no smallest supersets, the proposed model regards the largest subset corresponding to the given cue until time $\Delta T$ is regarded as the most similar description in $\Delta C$. In this case, the proposed model continues to observe the sensor data until its completion or until receiving new sensor data.

2. Uncertainty due to insufficient sensor data: If $\Delta T < T$ and $\Delta C$ do not have geometrical projection equivalent to the given description cue, then the model is uncertain at time $\Delta T$. The condition $\Delta T < T$ implies that all the pieces of description cue with regard to the action are not received. In this case, the proposed model adjusts its reference time in accordance to most similar pattern. The most similar pattern is chosen in the same way as in case 1. By adjusting the time reference, the proposed model is informed about the expected time of completion of the current activity considering received pieces of description. In this case, the received pieces of information are not sufficient to recognize the elderly and the proposed model continues to observe the next description cue with regard to current activity.

**Table 1.** Training sensor data of Activities of Daily Living dataset

|  | PIR ||| Magnetic |||| Flush | Pressure || Electric |||
|---|---|---|---|---|---|---|---|---|---|---|---|---|---|
|  | Shower | Basin | Cooktop | Main door | Fridge | Cabinet | Cupboard | Toilet | Seat | Bed | Microwave | Toaster | TV |
| Leaving | 0 | 0 | 0 | 1 | 0 | 0 | 0 | 0 | 0 | 0 | 0 | 0 | 0 |
| Toileting | 1 | 0 | 0 | 0 | 0 | 0 | 0 | 1 | 0 | 0 | 0 | 0 | 0 |
| Showering | 1 | 0 | 0 | 0 | 0 | 0 | 0 | 1 | 0 | 0 | 0 | 0 | 0 |
| Sleeping | 0 | 0 | 0 | 0 | 0 | 0 | 0 | 0 | 0 | 1 | 0 | 0 | 0 |
| Breakfast | 0 | 0 | 0 | 0 | 1 | 0 | 0 | 0 | 1 | 0 | 0 | 1 | 0 |
| Lunch | 0 | 0 | 1 | 0 | 1 | 0 | 0 | 0 | 0 | 0 | 1 | 0 | 0 |
| Dinner | 0 | 0 | 1 | 1 | 1 | 0 | 0 | 0 | 0 | 0 | 1 | 0 | 0 |
| Snacks | 0 | 0 | 0 | 0 | 1 | 0 | 0 | 0 | 1 | 0 | 1 | 0 | 0 |
| Spare_time | 0 | 0 | 0 | 1 | 0 | 0 | 0 | 0 | 1 | 0 | 0 | 0 | 1 |
| Grooming | 0 | 0 | 0 | 0 | 0 | 0 | 1 | 0 | 0 | 1 | 0 | 0 | 0 |

3. Uncertainty due no incoming sensor data: If $\Delta T = T$, $\Delta C$ do not have geometrical projection equivalent to the given description cue, then the model is uncertain at time $\Delta T$ and no piece of description cues received till estimated time of completion adjusted time reference indicates that wellness of the elderly has to ensured. Under this case, the proposed model alerts the kin to ensure the wellness of the elderly. Based on these three conditions, the proposed model assists elderly by identifying the situations based on its knowledge on which the kin of the elderly has to be alerted.

## 5      Experimental analysis

In this section of the paper, we have elaborated on the experimental analysis conducted on the proposed work using a sample data on human activity recognition. The dataset we have considered for this purpose is Activities of Daily Living dataset – A[22]. In this paper, we assume that the elderly we wish to assist lives in a smart home as in the dataset and the sensor data is centrally acquired by the proposed model via wireless link. The training sensor data lists the descriptions of 10 activities of the dataset in terms of 5 quality dimensions, 13 attributes and its relation as shown in Table 1. For this purpose, we have regarded the categories of sensors as quality dimensions, individual sensors under each category as attributes and their relation is represented with binary value [8]. The maximum of maximum time required for completion by the sample recordings corresponding to each activity in the dataset is the reference completion time for each activity as shown in Table 2.

The high dimensional conceptual space as a KID model learns the time driven activities in the terms of quality dimensions (sensors), attributes (types), and its relation (binary value) as geometrical structures. The proposed model stores the concepts obtained from KID model as geometrical structures in its memory as shown in Figure 2. It can be observed from Figure 2 that each concept is a geometrical structure consisting of projection formed by attributes under each quality dimension.

**Table 2.** Reference time of the proposed model

| Sleeping | Leaving | Toileting | Showering | Breakfast |
|---|---|---|---|---|
| **10:13:45 (T)** | 4:30:25 | 0:13:27 | 0:15:46 | 0:12:44 |
| Lunch | Dinner | Snacks | Spare time | Grooming |
| 0:35:56 | Not Available | 0:04:51 | 6:03:10 | 0:13:41 |

Upon learning the activities as concepts, the proposed model can now assists the elderly by continuously monitoring their activities in reference to its memory as mentioned in the previous section. Subsequently, experiments are conducted to check the activity recognizing ability of the proposed model. Please note that 24-hours format of time is followed in this section. As mentioned in the previous section, a total of three different cases are possible during activity recognizing. In the following, we have illustrated an example for each of the case.

Case 1- No uncertainty: Input from pressure sensor located in the bed is received at time 02:27 is received by the proposed model. The model detects the activity as sleeping. However, the reference time for sleeping activity is 10 hours 13 minutes and 45 seconds. This implies that the elderly is just settled to sleep. The proposed model continues to observe the sensor input till the reference time.

Case 2- Uncertainty due to insufficient sensor data: Input from magnetic sensor located in the fridge is received at time 10:34 is received by the proposed model. With this given information, the model is uncertain on the activity of the elderly. However, maximum reference time of activities involving fridge is 35 minutes and 56 seconds. The proposed model adjusts its reference time to 35 minutes and 56 seconds and continues to observe next sensor input. Similarly, the model receives successive inputs from magnetic sensor of the cupboard. With this current information, the model is still uncertain on the activity of the elderly. The proposed model continues observe the sensor data since it already has the maximum time reference of activities involving cupboard and fridge. Finally, an input from electric sensor of the toaster is received. With this information, the model is able to recognize the activity of the elderly is in wellness making his/ her breakfast.

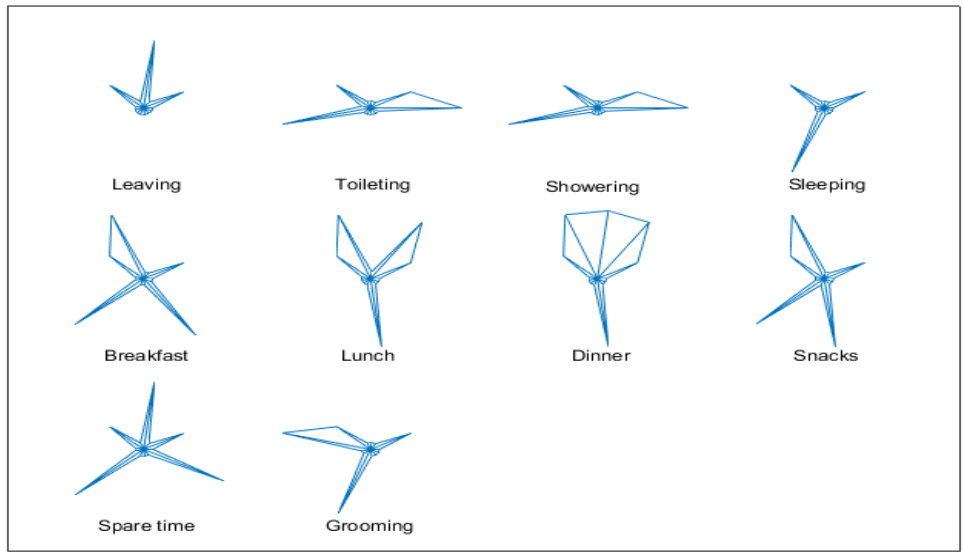

**Fig. 2.** Learning of activities by KID model as concepts

Case 3- Uncertainty due to no sensor data: There are no explicit examples for this case in the dataset. Let us provide an imaginary example for this case. Input from the magnetic sensor located in the main door is received at 13:09 is received by the proposed model. With this information, the model is certain that the elderly is leaving the house. The maximum reference time is adjusted from 10 hours 13 minutes and 45 seconds to 4 hours 30 minutes and 25 seconds. This is the maximum time the elderly as spent out as per the training data. If the elderly is not home by this time reference time, the kin is notified to check on the wellness of the elderly.

## 6    Discussion

In this paper, we have used the proposed cognitive system for elderly assistance by monitoring their daily activities. However, this proposed cognitive system can be slightly modified and can be used for other applications where anomaly (uncertainty) detection can be of much importance such as surveillance, system health monitoring, fault detection and so on. As mentioned above, the concept representation of high dimensional conceptual space is analogous to state representation of quantum theories. Similarly, projections formed by attributes of quality dimensions are analogous to basis vectors of quantum system. This derivation leaves us on an insight that slight refinement on geometrical framework of conceptual space lands us in relevance between conceptual space based cognition and quantum processes. On a straight forward note, geometry is the key for cognition in human brain, conceptual space and quantum theories. In this paper, we have shared our insights on quantum static information representation. However, dynamic quantum aspect on high dimensional conceptual space is one potential future work we wish to endeavor.

# 7 Conclusions

In this paper, we have proposed and modeled a cognitive system that assist elderly for their active living. The novelty of the cognitive system is adapting the proposed high dimensional conceptual space as KID model to acquire knowledge from the smart home sensor data and to recognize the activities of the elderly. Also, this cognitive system is capable of identifying and resolving constrained uncertainties. We have demonstrated via experimental analysis that adapting the proposed cognitive system can identify the uncertain situations on which wellness of the elderly has to be ensured.

**Acknowledgement:** This research has received financial support from Department of Science and Technology, Government of India under the scheme Cognitive Science Research Initiative with grant number: SR/CSIR/118/2014.